\documentclass{article}

\usepackage[numbers]{natbib}

\usepackage[final]{neurips_2024}

\usepackage[utf8]{inputenc} 
\usepackage[T1]{fontenc}    
\PassOptionsToPackage{hyphens}{url}
\usepackage{hyperref}       
\usepackage{url}            
\usepackage{booktabs}       
\usepackage{amsfonts}       
\usepackage{nicefrac}       
\usepackage{microtype}      
\usepackage[table,dvipsnames,svgnames]{xcolor}         
\usepackage{tabularx}       
\usepackage{pifont}
\usepackage{graphicx}
\usepackage{lipsum}
\usepackage{multirow}
\usepackage{marvosym}
\usepackage{xspace}
\usepackage{blindtext}
\usepackage{amsmath}
\usepackage{algorithm}

\usepackage{algpseudocode}
\usepackage{wrapfig}
\usepackage{enumitem}

\usepackage{subcaption}

\usepackage{amssymb}
\usepackage{amsmath}

\usepackage{tablefootnote}
\usepackage{subcaption}

\usepackage{wrapfig}
\usepackage{lipsum} 
\usepackage{booktabs} 
\usepackage{adjustbox}
\usepackage{afterpage}
\usepackage{float}
\usepackage{placeins}

\definecolor{catgray}{gray}{0.92}
\hypersetup{colorlinks,urlcolor=black,citecolor=cyan,linkcolor=purple}

\makeatletter
\DeclareRobustCommand\onedot{\futurelet\@let@token\@onedot}
\def\@onedot{\ifx\@let@token.\else.\null\fi\xspace}

\makeatother

\def\shortname{\mbox{PEvolve}\xspace}

\definecolor{citecolor}{rgb}{0.21,0.49,0.74}
\definecolor{linkcolor}{HTML}{ED1C24}
\definecolor{graycolor}{rgb}{0.95,0.95,0.95}
\definecolor{cvprblue}{rgb}{0.21,0.49,0.74}
\definecolor{memory}{rgb}{0.82, 0.51, 0.50} 
\definecolor{current}{rgb}{0.49, 0.60, 0.74} 

\usepackage[capitalize]{cleveref}
\crefname{section}{Sec.}{Secs.}

\crefname{table}{Tab.}{Tabs.}
\crefname{figure}{Fig.}{Figs.}

\title{ \shortname: Evolving Prompt Enhancer for Image Generation}

\author{
\textbf{PEvolve Team}
}
\begin{document}

\maketitle
\suppressfloats[t]

\begin{abstract}
Prompt enhancement rewrites user requests to improve image quality and adherence to user intent. Existing prompt
enhancers often rely on supervised fine-tuning on reference rewrites
or reinforcement learning with scalar rewards. However, these supervisory
signals provide limited explicit guidance on how rewriting strategies
affect generation outcomes or how to derive general rewriting strategies
from generation feedback across different requests. We introduce PEvolve,
a self-evolving prompt-enhancement framework that uses generation
feedback to evolve general rewriting skills while keeping all model
weights frozen. PEvolve incorporates two key components. Hierarchical
Comparative Reflection builds on image-level analyses to combine
within-group comparison of multiple attempts for the same request
with cross-group induction of general strategy updates. Weakness
Exploration constructs targeted probes for each candidate skill edit
and refines it using generation feedback before validating the updated
library. Experiments on infographic, general-object, and
knowledge-reasoning tasks show consistent improvements over prompt
enhancement with the initial skill library. Further experiments across
prompt-enhancement models and image generators also show consistent
gains, including configurations with built-in prompt enhancement
enabled, supporting the framework's broad adaptability.
\end{abstract}

\makeatletter
\def\@makefnmark{}
\makeatother

\section{Introduction}

\begin{figure}[t]
    \centering
    \includegraphics[width=\textwidth]{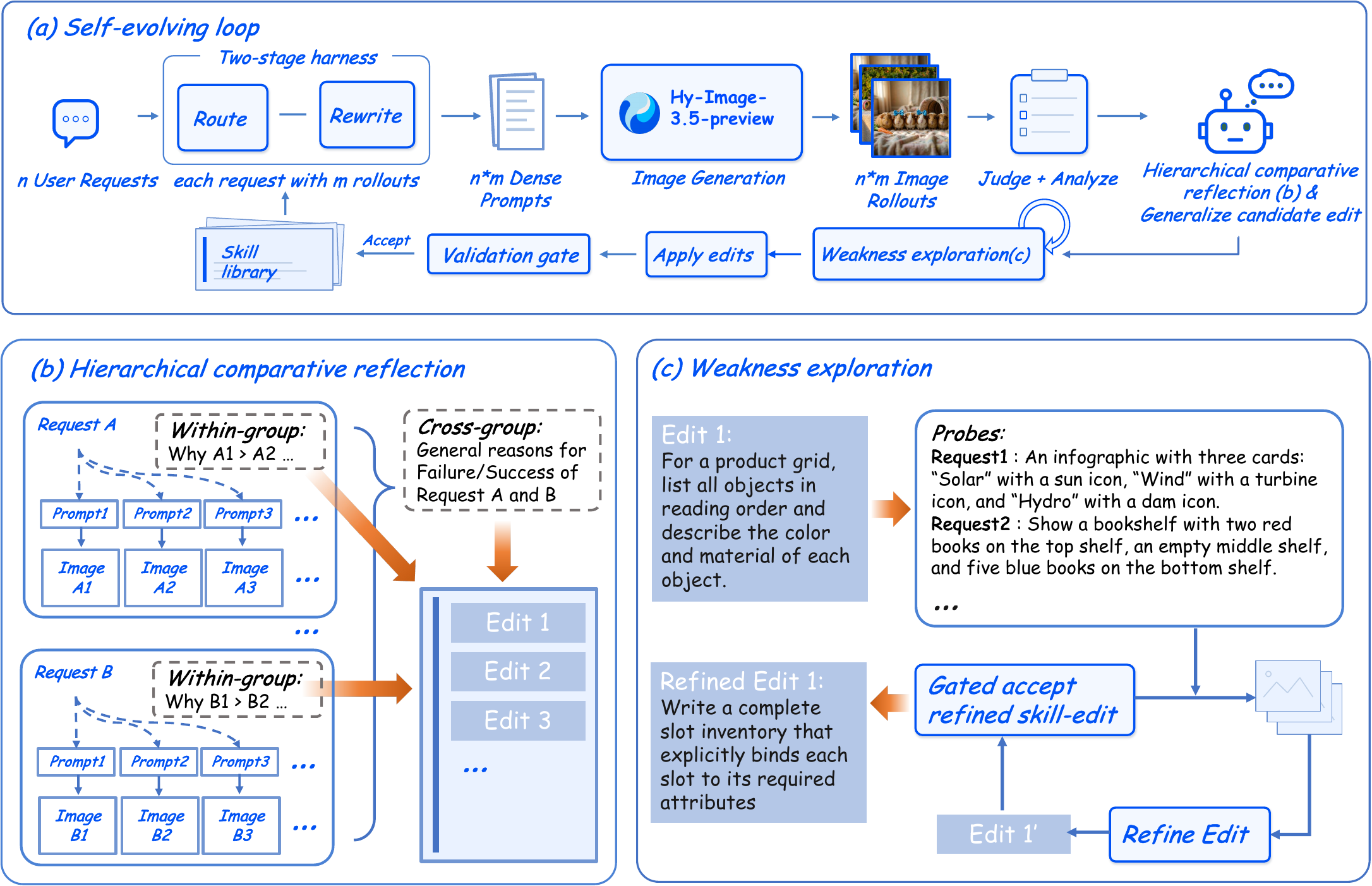}
    \caption{Overview of PEvolve. (a) The skill library evolves through generation, reflection, edit refinement, and validation, while all model weights remain frozen. (b) Each generation attempt is inspected independently and summarized into textual evidence. Type-aware batches organize request groups for within-group comparison and cross-group induction. (c) Targeted probes provide feedback for refining candidate skill edits before the resulting library is validated.}
    \label{fig:pevolve_framework}
\end{figure}

Producing images that faithfully reflect user intent depends not only on the capabilities of a text-to-image model, but also on how the intended content is expressed. User requests are often concise or underspecified, while image generators can be sensitive to the wording, structure, and level of detail in their inputs~\cite{wang2023reprompt}. Prompt enhancement (PE) rewrites user requests into prompts better suited to the target generator while preserving the user's intent~\cite{betker2023improving}. However, effective enhancement requires understanding both the user's intent and the target generator's capabilities and limitations. Even a linguistically clear and complete description may demand complex relationships, precise counting, or text rendering that the generator struggles to realize~\cite{huang2023t2i,ghosh2023geneval,tuo2024anytext}, and adding detail does not necessarily resolve these difficulties. PE should therefore go beyond making prompts longer or more descriptive, accounting for the generator's strengths and limitations to better realize the intended content.

Existing prompt enhancers typically rely on reference rewrites for supervised fine-tuning (SFT) or image-based scalar rewards for reinforcement learning (RL)~\cite{wu2026reprompt}. SFT is susceptible to training-data biases: enhancers may inherit the content distributions, expression preferences, and elaboration patterns of reference rewrites. These patterns may not suit new user requests or the target generator, potentially introducing unsupported details, altering user intent, or making generation more difficult. RL optimizes enhancers using actual generation outcomes, but scalar rewards provide limited insight into the reasons behind a score. The same low score may reflect omitted constraints, ambiguous wording, or limited generator capabilities, each requiring different strategy adjustments. Imperfect rewards may also be overoptimized, creating a disconnect between higher scores and improvements in actual quality~\cite{gao2023scaling}. We therefore need to identify the effects and limitations of rewriting strategies from generation feedback and turn these observations into general strategies that apply across requests.

In this report, we propose \textbf{PEvolve}, a framework that evolves general rewriting skills through a self-evolving loop driven by generation feedback. The skills encode explicit rewriting strategies that the enhancer retrieves and applies to user requests. During optimization, PEvolve samples multiple enhanced prompts for each request and generates corresponding images, forming a group of generation attempts grounded in the same user intent. Based on the evaluations and analyses of these attempts, the optimizer uses two parallel and complementary reflection strategies. \textbf{Within-group comparison} examines performance differences among attempts for the same request to extract effective rewriting strategies and identify persistent difficulties. \textbf{Cross-group induction} analyzes generation outcomes across requests to identify common reasons for success and failure and derive general rewriting rules with appropriate applicability conditions. Both strategies can propose candidate skill edits. All model weights remain frozen throughout this process, while updates to explicit skills accumulate experience that guides prompt enhancement for subsequent requests.

However, rules inferred from a limited set of generation outcomes may fit their motivating examples without adequately capturing the conditions under which they fail. To address this issue, PEvolve incorporates \textbf{Weakness Exploration} to probe and refine candidate skill edits before incorporating them into the library. For each proposed edit, it constructs targeted test cases that examine the rule beyond the observations from which it was derived, and uses the resulting generation feedback in a local refinement loop. This process seeks to uncover overly narrow instructions, inappropriate generalizations, and unintended effects, providing concrete evidence for revising the rule's content and applicability conditions. The refined edits are subsequently assembled into a candidate library for validation.

Our main contributions are as follows:
\begin{itemize}
    \item We introduce PEvolve, a framework that evolves general rewriting skills through generation feedback without updating model weights. Hierarchical Comparative Reflection combines within-group comparison and cross-group induction to propose candidate skill edits, while Weakness Exploration probes and refines these edits.

    \item We evaluate PEvolve on infographic generation, general-object generation, and knowledge reasoning, observing improvements over prompt enhancement with the initial skill library on all three tasks. We also use out-of-domain GSB comparisons to assess the applicability of trained skills to requests beyond the training domains.

    \item We conduct systematic cross-backbone evaluations with multiple prompt-enhancement LLMs and image generators. These experiments establish the framework's applicability across the evaluated model configurations and show that its benefits are not confined to a particular enhancer--generator pairing.
\end{itemize}

\begin{figure}[t]
        \centering
        \includegraphics[width=\textwidth]{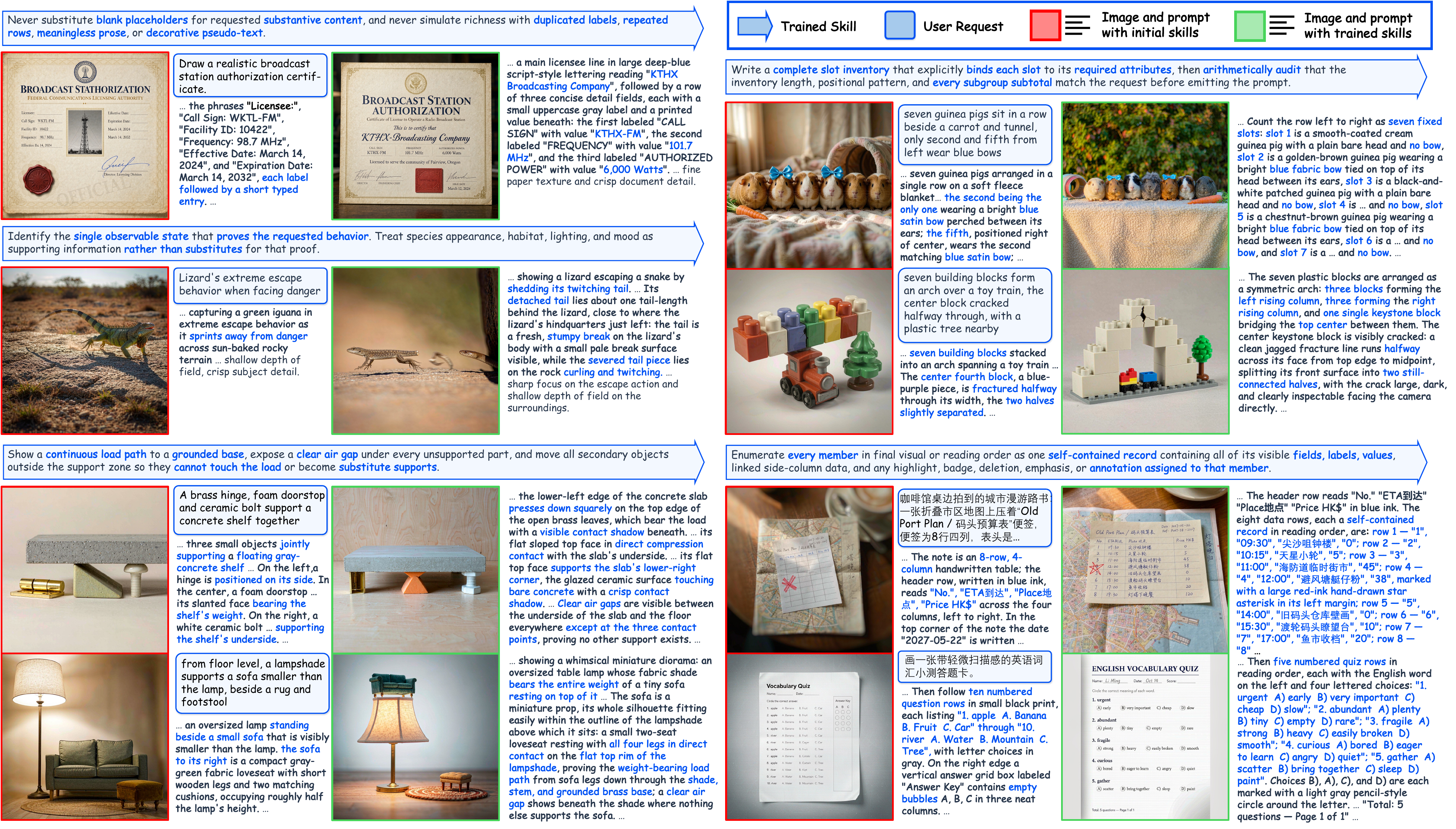}
        \caption{
            PEvolve evolves general rewriting skills from
            generation feedback without updating model weights.
            Each example shows a user request, rewritten prompt
            excerpts, and generated images using the initial skill
            library (red borders) and the trained skill library
            (green borders). Blue arrows show excerpts from trained
            skills, with shared arrows illustrating how the same
            rule guides different requests. Blue text highlights
            rule-relevant prompt content, and ellipses indicate
            omitted text.
        }
        \label{fig:teaser}
    \end{figure}

\section{Related Work}

\paragraph{Prompt Enhancer}
Prompt enhancers translate user requests into formulations better suited to text-to-image generators.
Promptist~\cite{hao2023optimizing} and PRIP~\cite{zhan2024prompt} combine supervised learning with preference-driven optimization, while PromptEnhancer~\cite{wang2025promptenhancer} trains chain-of-thought rewriting with fine-grained image--text alignment rewards.
APE~\cite{huang2026ape} post-trains small language models with task-aware rewards for image generation and editing, supporting both single-pass rewriting and a multi-agent router--rewriter--composer pipeline.
Visual feedback also informs supervision beyond scalar scoring: FaithRewriter~\cite{liu2026seeing} constructs visually verified revisions for supervised and preference learning, and PRISM~\cite{tang2026prism} learns image-grounded diagnosis and feedback-conditioned rewriting alongside reward-based optimization.
Complementary approaches optimize prompts at test time without updating model weights.
VisualPrompter~\cite{wu2026visualprompter} checks atomic semantic requirements in generated images and revises prompts to address missing concepts, whereas RATTPO~\cite{kim2025reward} summarizes prompt--reward histories into textual optimization hints for each request.
These studies demonstrate that generation feedback can support both learned rewriting policies and request-specific refinement.
PEvolve instead uses such feedback to explicitly revise general rewriting skills, so that observations from generation attempts inform the rewriting strategies applied to subsequent requests rather than only the current prompt or the rewriter's parameters.

\paragraph{Self-evolving Strategy}
Self-evolving agents can accumulate experience through external instructions and skills rather than parameter updates~\cite{shinn2023reflexion,zhao2024expel,wang2023voyager}.
TextGrad~\cite{yuksekgonul2024textgrad} propagates natural-language feedback to optimize textual variables, and GEPA~\cite{agrawal2026gepa} evolves prompts through reflection on execution traces and candidate evaluation.
ACE~\cite{zhang2026agentic} organizes experience into an evolving playbook, while SkillOpt~\cite{yang2026skillopt} and SkillOpt-Lite~\cite{shen2026skillopt} revise skill documents from execution evidence and retain updates through validation.
SkillOpt evaluates document visual question answering, and XSkill~\cite{jiang2026xskill} combines within-task cross-rollout critique with experience and skill consolidation for multimodal tool use.
Related mechanisms have also been explored in visual generation.
MemoGen~\cite{chen2026memogen} retrieves polarity-labeled, instance-level generation memories with frozen models; VideoWeaver~\cite{wei2026videoweaver} evolves long-video composition and creator skills; and GenEvolve~\cite{chen2026genevolve} extracts visual experience from same-request trajectory comparisons and distills it into agent parameters.
The Auto Skill Evolver (ASE) introduced with D-OPCD~\cite{wang2026internalizing} consolidates insights from generation trajectories into skills that guide the construction of initial generation prompts. D-OPCD then uses agent-improved prompts as privileged context to distill these capabilities into the diffusion generator.
Within this broader direction, PEvolve focuses on evolving the general rewriting skills of a prompt enhancer.
It combines within-group comparison of alternative rewrites and generated images with cross-group induction of rewriting rules, and uses Weakness Exploration to test and refine candidate skill edits before library-level validation.
Throughout this process, generation feedback updates explicit prompt-enhancement strategies while all model weights remain frozen.

\section{PEvolve}
\subsection{Framework}

\paragraph{Self-Evolving Loop}
As illustrated in Figure~\ref{fig:pevolve_framework}(a), PEvolve improves a prompt enhancer by iteratively updating the general rewriting skills in its library while keeping all model weights frozen. Let $\mathcal{L}_t$ denote the library at iteration $t$, and let $P$, $G$, $J$, and $O$ denote the prompt enhancer, image generator, judge, and optimizer, respectively. Prompt enhancement follows a two-stage harness: a router selects relevant skills based on the original request and skill descriptions, and the prompt enhancer then rewrites the request using the selected skill contents. Starting from an initial library $\mathcal{L}_0$, each iteration collects generation experience, proposes and refines candidate skill edits, and uses validation to decide whether to update the library. For a batch of requests $\mathcal{B}_t=\{q_i\}_{i=1}^{n}$, we perform $m$ generation attempts per request, each following routing, rewriting, image generation, and evaluation:
\begin{equation}
\begin{aligned}
\mathcal{I}_{ij} &= \operatorname{Route}(q_i,\operatorname{Desc}(\mathcal{L}_t)), \\
p_{ij} &\sim P(\cdot\mid q_i,\mathcal{L}_t[\mathcal{I}_{ij}]), \\
x_{ij} &\sim G(\cdot\mid p_{ij}), \qquad (s_{ij},f_{ij})=J(q_i,x_{ij}),
\end{aligned}
\label{eq:pevolve_rollout}
\end{equation}
where $q_i$ is the original user request, $p_{ij}$ is its $j$-th enhanced prompt, and $x_{ij}$ is the corresponding generated image, with $j\in\{1,\ldots,m\}$. $\operatorname{Desc}(\mathcal{L}_t)$ contains the skill identifiers and descriptions, $\mathcal{I}_{ij}$ is the set of skill identifiers selected for this attempt, and $\mathcal{L}_t[\mathcal{I}_{ij}]$ denotes the corresponding skills used for rewriting. The judge evaluates $x_{ij}$ against the original request $q_i$, producing a scalar score $s_{ij}$ and an accompanying assessment $f_{ij}$. The $m$ attempts associated with $q_i$ form a request group.

We then employ \emph{Hierarchical Comparative Reflection} (Figure~\ref{fig:pevolve_framework}(b)) to derive candidate skill edits through within-group comparison and cross-group induction. To accommodate the limited visual context of VLMs, we decouple case-level visual inspection from batch-level reflection. The optimizer first examines one generation attempt at a time:
\begin{equation}
a_{ij}=O_{\mathrm{vis}}(q_i,p_{ij},x_{ij},s_{ij},f_{ij},\mathcal{I}_{ij}),
\label{eq:pevolve_analysis}
\end{equation}
where $a_{ij}$ is a concise textual analysis of the visual outcome and its implications for rewriting. The selected skill identifiers $\mathcal{I}_{ij}$ accompany each attempt throughout analysis and reflection. Together with the current library, they allow the optimizer to relate generation outcomes to the skills selected for rewriting and propose targeted changes to skill contents or routing descriptions. The optimizer subsequently reflects on these textual records in minibatches and consolidates their proposals, preserving both request-group identities and per-attempt skill selections. Here, $O_{\mathrm{vis}}$ and the subsequent text-based operations are stages of the same optimizer.

We further use type-aware batching, preferentially grouping requests of the same task type for reflection while varying the specific instances. Such requests are more likely to share semantic constraints and generation difficulties. Grouping them helps the optimizer identify shared rewriting issues and derive rules that apply across instances. After reflection, aggregation, and selection, the textual evidence yields a bounded set of candidate edits:
\begin{equation}
\Delta_t=O_{\mathrm{text}}\!\left(\mathcal{L}_t,\{(q_i,\mathcal{I}_{ij},p_{ij},s_{ij},f_{ij},a_{ij})\}_{i,j}\right),
\label{eq:pevolve_proposal}
\end{equation}
where $O_{\mathrm{text}}$ summarizes reflection, aggregation, and selection, using the recorded skill identifiers to link each attempt to the corresponding entries in $\mathcal{L}_t$.

The candidate edits then undergo \emph{Weakness Exploration} (Figure~\ref{fig:pevolve_framework}(c)), which uses targeted cases to test and refine the proposed rules, producing refined edits $\widehat{\Delta}_t$. Applying them gives a candidate library $\widehat{\mathcal{L}}_t=\operatorname{Apply}(\mathcal{L}_t,\widehat{\Delta}_t)$. We evaluate this library on a fixed held-out validation set. SkillOpt~\cite{yang2026skillopt} accepts a candidate only if its validation score strictly exceeds that of the current skill. In our setting, stochasticity in both prompt rewriting and image generation can obscure small gains or yield an unusually high incumbent score. Requiring every update to surpass that score may reject potentially useful incremental edits before their benefits can accumulate. We therefore set the acceptance threshold to the mean of the best and second-best historical validation scores:
\begin{equation}
\begin{aligned}
\tau_t &= \frac{v_t^{(1)}+v_t^{(2)}}{2}, \\
\mathcal{L}_{t+1} &= \begin{cases}
\widehat{\mathcal{L}}_t, & \operatorname{Val}(\widehat{\mathcal{L}}_t)>\tau_t, \\
\mathcal{L}_t, & \text{otherwise}.
\end{cases}
\end{aligned}
\label{eq:pevolve_gate}
\end{equation}
Here, $\operatorname{Val}$ is the mean validation score. The values $v_t^{(1)}$ and $v_t^{(2)}$ are the highest and second-highest distinct validation scores observed before evaluating the current candidate. Both are initialized to $\operatorname{Val}(\mathcal{L}_0)$. This gate allows high-performing intermediate libraries below the historical best to guide subsequent iterations. The next iteration collects experience using $\mathcal{L}_{t+1}$, while the best-scoring library is retained separately for deployment. At inference time, the enhancer retrieves and applies its skills.

\paragraph{Hierarchical Comparative Reflection}
We combine within-group comparison and cross-group induction as two parallel and complementary strategies (Figure~\ref{fig:pevolve_framework}(b)) to derive general rewriting strategies from generation feedback and propose candidate skill edits. The optimizer receives the current skill library and textual records organized by user request, including rewritten prompts, selected skill identifiers, judge scores and assessments, and per-attempt visual analyses.

In within-group comparison, the optimizer uses judge results and visual analyses to compare alternative rewrites and their selected skills for the same request. It examines the effective expressions used in better-performing attempts and the requirements omitted or misinterpreted in worse-performing attempts. Difficulties that persist across attempts also provide clues to the limitations of rewriting strategies and the capabilities of the generator.

In cross-group induction, the optimizer analyzes generation outcomes across requests to identify common reasons for success and failure, as well as shared effective strategies and generation difficulties. Using the expressions, skill selections, and generation feedback from these requests, it further examines whether the difficulties stem from shortcomings in rewriting strategies or limitations of the generator. It derives general rewriting rules that accommodate these limitations, specifies their applicability conditions, and uses them to revise skill contents or routing descriptions. Together, the two strategies provide candidate edits for skill-library evolution, which are subsequently tested and refined through Weakness Exploration.

\paragraph{Weakness Exploration}
Rules induced from a finite set of generation attempts may be constrained by their originating cases and fail to cover other situations governed by the same principle. We therefore introduce a probe-and-refine loop for candidate skill edits to examine and improve the generality of the rules before updating the skill library (Figure~\ref{fig:pevolve_framework}(c)). For each candidate edit $e\in\Delta_t$ to be refined, the optimizer uses the edit and its originating training context to construct probe requests $\mathcal{Q}_e$ that span different domains and trigger modes while sharing the general principle underlying the rule. The probe requests for each edit remain fixed across refinement rounds.

At refinement round $r$, we apply only the current edit version $e^{(r)}$ to the library at the start of iteration $t$:
\begin{equation}
\mathcal{L}_{t,e}^{(r)}=
\operatorname{Apply}(\mathcal{L}_t,\{e^{(r)}\}),
\label{eq:pevolve_edit_probe}
\end{equation}
where $e^{(0)}=e$. Each request in $\mathcal{Q}_e$ follows the same routing, prompt rewriting, image generation, judging, and per-attempt analysis pipeline as the main loop. Let $\bar{s}_e^{(r)}$ denote the mean judge score across all generation attempts on $\mathcal{Q}_e$. Based on the probe outcomes and visual analyses, the optimizer adjusts the wording of the rule's trigger conditions or mechanism to improve its applicability across situations while preserving its intended objective, operation type, and effective behavior.

Refinement stops when the probe score reaches an early-stopping threshold or the preset limit on evaluation rounds is reached. We retain the tested version with the highest mean judge score:
\begin{equation}
 r_e^{\star}=\arg\max_{0\leq r\leq R_e}\bar{s}_e^{(r)},
 \qquad \widehat{e}=e^{(r_e^{\star})},
\label{eq:pevolve_edit_selection}
\end{equation}
where $R_e$ is the final evaluated round. The refined edits are combined with any edits not subject to refinement to form $\widehat{\Delta}_t$. Applying these edits to the current skill library $\mathcal{L}_t$ yields a candidate library. We then evaluate the joint effect of the edits on the validation set to decide whether to accept the update.

\subsection{Training Recipe}

\paragraph{Task-wise Setup}
Our main experiments cover knowledge reasoning, infographic generation, and general-object generation. Each task uses 200 training requests, 200 validation requests, and 600 test requests, and independently evolves a skill library from the same initialization $\mathcal{L}_0$ under a shared optimization configuration. Data construction and splits are detailed in Appendix~\ref{app:data_construction}. The sources and construction of the initial skill library are described in Appendix~\ref{app:initial_skill_library}. We use GLM-5.3-Flash~\cite{glm5team2026glm5} as the prompt enhancer, GPT-5.6 Sol~\cite{openai2026gpt56} as the optimizer, Gemini 3.1 Pro Preview~\cite{deepmind2026gemini31pro} as the judge, and Hy-Image-3.5-preview~\cite{tencent2026hyimage35} as the image generator. All model weights remain frozen. Training requests supply generation experience, validation requests support update acceptance and library selection, and test requests are reserved for final evaluation.

\paragraph{Task-specific Evaluation}
We evaluate each task using its corresponding scoring dimensions. For knowledge reasoning, the judge assigns a binary score according to whether the image correctly realizes the knowledge-dependent answer implied by the original request. The average score gives task accuracy. Infographic generation is evaluated on prompt consistency, text accuracy, layout and hierarchy, visual aesthetics, and information richness. General-object generation is evaluated on core constraint correctness, object and relation completeness, and prompt adherence and legibility. For the latter two tasks, each dimension receives a continuous score in $[0,1]$, and the equally weighted mean of the dimension scores supplies the scalar score used during optimization and validation. Final evaluation reports each task's corresponding metrics averaged over its test requests.

\paragraph{Optimization and Selection}
Each iteration processes a batch of 40 requests with four generation attempts per request. Reflection uses minibatches with a target size of 20 requests, preferentially grouping requests of the same type. The router selects at most six skills per attempt. The optimizer can append, insert, replace, or delete skill content, revise routing descriptions, and add, remove, or merge skills. The per-iteration candidate skill edit budget follows a cosine schedule from four to two edits and remains at two if training extends beyond the nominal schedule. For each edit subject to Weakness Exploration, the optimizer generates 20 dedicated probe requests and evaluates four generation attempts per probe. We allow at most two evaluation rounds per edit, including the initial candidate evaluation, and stop refinement early if the mean probe score reaches 0.9. We retain the tested version with the highest mean probe score.

Validation and test evaluation use one generation attempt per request. We apply the best/second-best mean gate in Equation~\ref{eq:pevolve_gate}. Training stops only when at least two epochs have been completed and at least three candidate-library updates have passed this gate; otherwise, additional iterations are performed. Each accepted candidate library contributes one to this count. The validation-best library is retained separately from the current library and used for the corresponding task's final evaluation.

\paragraph{Skill Composition for OOD Evaluation}
For out-of-domain (OOD) evaluation, we combine the validation-best libraries from the three tasks. Non-conflicting skill updates are adopted directly, while conflicting versions are merged by GPT-5.6 Sol, which is instructed to preserve the original wording as much as possible while reconciling the conflicts. Skills unchanged across tasks retain their initial versions. Our human evaluation covers 400 image pairs generated for OOD requests that approximate the distribution of real-world user requests and cover domains outside the training tasks. The comparison uses the initial library and the merged trained library with the same GLM-5.3-Flash enhancer, Hy-Image-3.5-preview generator, and generation settings. Human annotators perform a Good/Same/Bad (GSB) comparison, judging whether the trained-library output is better than, equivalent to, or worse than the initial-library output.

\section{Experiments}

\begin{table*}[t]
\centering
\small
\renewcommand{\arraystretch}{1.15}
\caption{Main results across three image-generation tasks. All scores are multiplied by 100. Higher is better. Panel (a) reports overall performance, with parenthesized values indicating PEvolve's absolute gains over Initial Skills. Panel (b) reports individual evaluation dimensions.}
\label{tab:main_results}

\begin{tabularx}{\textwidth}{@{}l*{3}{>{\raggedright\arraybackslash}X}@{}}
\toprule
\multicolumn{4}{@{}l}{\textbf{(a) Overall performance}} \\
\addlinespace[3pt]
Method & Infographic & General objects & Knowledge reasoning \\
\midrule
No PE & 64.81 & 48.43 & 33.00 \\
Initial Skills & 80.71 & 71.50 & 66.67 \\
\rowcolor{gray!12}
\rowcolor{gray!12}
\textbf{PEvolve}
& \mbox{\textbf{88.76}\,{\scriptsize(+8.05)}}
& \mbox{\textbf{75.64}\,{\scriptsize(+4.14)}}
& \mbox{\textbf{72.17}\,{\scriptsize(+5.50)}} \\
\bottomrule
\end{tabularx}

\vspace{7pt}
\begin{tabularx}{\textwidth}{@{}l*{8}{>{\centering\arraybackslash}X}@{}}
\toprule
\multicolumn{9}{@{}l}{\textbf{(b) Dimension-level results}} \\
\addlinespace[3pt]
& \multicolumn{5}{c}{Infographic}
& \multicolumn{3}{c}{General objects} \\
\cmidrule(lr){2-6}\cmidrule(l){7-9}
Method & PC & TA & LH & VA & IR & CC & OR & AL \\
\midrule
No PE & 78.04 & 43.68 & 83.65 & 83.08 & 35.59 & 32.52 & 53.45 & 59.32 \\
Initial Skills & 90.48 & 70.71 & 91.59 & 91.36 & 59.41 & 56.06 & 77.86 & 80.58 \\
\rowcolor{gray!12}
\textbf{PEvolve}
& \textbf{94.65} & \textbf{79.57} & \textbf{94.73}
& \textbf{94.34} & \textbf{80.49} & \textbf{62.85}
& \textbf{80.72} & \textbf{83.35} \\
\bottomrule
\end{tabularx}

\par\smallskip
\begin{minipage}{\textwidth}
\footnotesize
PC: prompt consistency; TA: text accuracy; LH: layout and hierarchy;
VA: visual aesthetics; IR: information richness;
CC: core-constraint correctness; OR: object and relation completeness;
AL: prompt adherence and legibility.
Infographic and general-object overall scores average their respective
dimensions. Knowledge reasoning is evaluated by accuracy.
\end{minipage}
\end{table*}

\subsection{Main Results}
We compare three configurations: direct image generation from the original requests (No PE), prompt enhancement with the initial skill library (Initial Skills), and prompt enhancement with the trained skill library (PEvolve). The latter two use the same two-stage routing and rewriting harness and differ in the skill library they use. All three configurations are evaluated on the same 600 test requests per task, with one generated image per request. PEvolve uses the library with the highest validation score during training for final testing.

As shown in Table~\ref{tab:main_results}, Initial Skills outperforms No PE on all three tasks, showing that prompt enhancement with the initial skill library already improves generation performance. With the harness and models held fixed, PEvolve further improves performance on all tasks by evolving the skill library: infographic and general-object scores increase by 8.05 and 4.14 points, respectively, and knowledge-reasoning accuracy increases by 5.50 percentage points. These results show that generation feedback can guide continued improvement of the skill library, enabling the trained rewriting strategies to produce better images for held-out test requests.

The dimension-level results show improvements across all five infographic dimensions and all three general-object dimensions. For infographics, information richness shows the largest gain, increasing from 59.41 to 80.49, while text accuracy rises from 70.71 to 79.57. For general objects, the largest improvement is in core-constraint correctness, from 56.06 to 62.85. Overall, skill evolution improves multiple aspects of generation, including content expression and constraint satisfaction. Qualitative comparisons for the three tasks are provided in Appendix~\ref{app:qualitative_cases}.

\paragraph{Human Evaluation on Out-of-Domain Requests.}
To assess the applicability of trained skills beyond the training domains, we compare the library composed from the three task-specific trained libraries with the initial library on OOD requests that approximate the distribution of real-world user requests, using human GSB evaluation of 400 image pairs. As shown in Figure~\ref{fig:gsb_human_evaluation} in Appendix~\ref{app:evaluation_details}, PEvolve is judged better in 56\% of comparisons, equivalent in 14\%, and worse in 30\%, with the proportion of better outcomes exceeding that of worse outcomes by 26 percentage points. These results indicate that rewriting experience acquired from the training tasks can transfer to OOD requests from unseen domains, providing further support for the generality of the trained skills.

\subsection{Ablations}
\label{sec:ablations}

\begin{table*}[t]
\centering
\caption{Adaptation across image generators and skill initializations. Scores are scaled by 100. Higher is better. The lower block uses initial skills obtained by decomposing Qwen-Image-2512's official PE prompt.}
\label{tab:generator_adaptation}
\begingroup
\small
\setlength{\tabcolsep}{2pt}
\renewcommand{\arraystretch}{1.10}
\begin{tabular*}{\textwidth}{@{\extracolsep{\fill}}lllrrr@{}}
\toprule
Generator & Built-in PE & Method
& Infographic & General objects & Knowledge reasoning \\
\midrule
Hy-Image-3.5-preview (main) & off & Initial Skills
& 80.71 & 71.50 & 66.67 \\
& & PEvolve
& \textbf{88.76} & \textbf{75.64} & \textbf{72.17} \\
\addlinespace[3pt]
Qwen-Image-2.0 & on & Initial Skills
& 85.40 & 64.00 & 65.10 \\
& & PEvolve
& \textbf{90.10} & \textbf{67.80} & \textbf{66.50} \\
\addlinespace[3pt]
Qwen-Image-2.0 & off & Initial Skills
& 84.40 & 67.00 & 67.20 \\
& & PEvolve
& \textbf{88.80} & \textbf{70.30} & \textbf{70.20} \\
\addlinespace[3pt]
Qwen-Image-2512 & off & Initial Skills
& 79.00 & 61.90 & 64.80 \\
& & PEvolve
& \textbf{81.40} & \textbf{64.70} & \textbf{69.60} \\
\addlinespace[3pt]
Seedream 5.0 Pro & off & Initial Skills
& 92.40 & 79.50 & 72.40 \\
& & PEvolve
& \textbf{95.20} & \textbf{85.80} & \textbf{75.70} \\
\midrule
\multicolumn{6}{l}{\textit{Initialization from Qwen-Image-2512's official PE prompt}} \\
Qwen-Image-2512 & off & Initial Skills
& 79.00 & 56.70 & 63.20 \\
& & PEvolve
& \textbf{80.40} & \textbf{61.90} & \textbf{65.20} \\
\bottomrule
\end{tabular*}
\endgroup
\end{table*}

\paragraph{Adaptation across image generators.}
We keep the PE model, optimizer, and judge fixed and separately evolve the skill library for each generator and task. As shown in Table~\ref{tab:generator_adaptation}, PEvolve improves performance on all three tasks with Qwen-Image-2.0~\cite{zhao2026qwenimage2}, Qwen-Image-2512~\cite{wu2025qwenimage}, and Seedream 5.0 Pro~\cite{bytedance2026seedream5pro}. For Qwen-Image-2.0, skill evolution yields gains with built-in prompt enhancement both enabled and disabled, indicating that trained skills can complement the generator's existing prompt enhancement capabilities. Seedream 5.0 Pro has the strongest initial performance among the compared generators, yet its general-object score still increases from 79.50 to 85.80. We further construct an alternative initial skill library by decomposing Qwen-Image-2512's official PE prompt into skills. All three tasks also improve under this initialization, supporting PEvolve's adaptability to different skill initializations.

Skills trained for different generators exhibit different strategies for handling the same types of constraints. For exact counting and state binding, skills trained for Seedream 5.0 Pro emphasize enumerating objects individually and explicitly assigning their states. Skills trained for Qwen-Image-2.0 with built-in PE disabled instead favor partitioning objects into smaller groups, binding states within each group, and specifying group subtotals. The former makes the correspondence between counts and attributes explicit through descriptions of individual objects, while the latter organizes local states and overall count constraints through grouping and subtotals. Together with the performance gains across generators, these strategy differences indicate that PEvolve can use generation feedback to adjust how constraints are expressed, adapting its rewriting strategies to the capabilities of different generators.

\begin{table*}[t]
\centering
\caption{Adaptation across PE models with Hy-Image-3.5-preview as the image generator. Scores are scaled by 100. Higher is better. The optimizer, judge, and initial skill library are held fixed, and each PE model is evaluated on the same test data. GLM-5.3-Flash results are reproduced from the main experiments for reference.}
\label{tab:pe_adaptation}
\begingroup
\small
\setlength{\tabcolsep}{6pt}
\renewcommand{\arraystretch}{1.10}
\begin{tabular*}{\textwidth}{@{\extracolsep{\fill}}llrrr@{}}
\toprule
PE model & Method & Infographic & General objects
& Knowledge reasoning \\
\midrule
GLM-5.3-Flash (main) & Initial Skills
& 80.71 & 71.50 & 66.67 \\
& PEvolve
& \textbf{88.76} & \textbf{75.64} & \textbf{72.17} \\
\addlinespace[3pt]
Hy3 & Initial Skills
& 77.39 & 68.73 & 58.00 \\
& PEvolve
& \textbf{84.07} & \textbf{72.06} & \textbf{62.83} \\
\addlinespace[3pt]
Qwen3.8-Flash & Initial Skills
& 74.91 & 68.21 & 65.17 \\
& PEvolve
& \textbf{78.17} & \textbf{71.90} & \textbf{68.33} \\
\addlinespace[3pt]
GPT-5.6 Sol & Initial Skills
& 87.89 & 74.56 & 71.67 \\
& PEvolve
& \textbf{93.17} & \textbf{79.02} & \textbf{73.83} \\
\bottomrule
\end{tabular*}
\endgroup
\end{table*}

\paragraph{Adaptation across PE models.}
We keep the image generator (Hy-Image-3.5-preview), optimizer, and judge fixed while varying the PE model, independently evolving the same initial skill library for each PE model and task. As shown in Table~\ref{tab:pe_adaptation}, PEvolve improves performance on all three tasks with GLM-5.3-Flash, Hy3~\cite{tencent2026hy3}, Qwen3.8-Flash~\cite{alibaba2026qwen38flash}, and GPT-5.6 Sol. GPT-5.6 Sol has the strongest initial performance on all three tasks, yet skill evolution further increases its infographic score from 87.89 to 93.17, its general-object score from 74.56 to 79.02, and its knowledge-reasoning accuracy from 71.67\% to 73.83\%. These results show that PEvolve can improve rewriting performance across PE models through skill evolution, complementing their existing enhancement capabilities and supporting the framework's adaptability to different PE models.

\paragraph{Ablation of Core Components.}
We separately remove Hierarchical Comparative Reflection and Weakness Exploration to examine the contribution of each component to skill evolution. Table~\ref{tab:core_ablation} compares the full PEvolve framework with these two variants across the three tasks.

\begin{table}[H]
\centering
\small
\renewcommand{\arraystretch}{1.15}
\caption{Ablation of the two core components. Scores are scaled by 100. Higher is better.}
\label{tab:core_ablation}
\setlength{\tabcolsep}{3pt}
\begin{tabularx}{\textwidth}{@{}l>{\hsize=.75\hsize\centering\arraybackslash}X>{\hsize=\hsize\centering\arraybackslash}X>{\hsize=1.25\hsize\centering\arraybackslash}X@{}}
\toprule
Method & Infographic & General objects & Knowledge reasoning \\
\midrule
\rowcolor{gray!12}
\textbf{PEvolve} & 88.76 & 75.64 & 72.17 \\
w/o Hierarchical Comparative Reflection & 86.11 & 73.15 & 70.17 \\
w/o Weakness Exploration & 85.15 & 75.38 & 70.67 \\
\bottomrule
\end{tabularx}
\end{table}

Removing either component lowers performance on all three tasks. Removing Hierarchical Comparative Reflection leads to larger drops in general-object generation and knowledge reasoning, whereas removing Weakness Exploration causes the larger drop in infographic generation.

\section{Conclusion}
We presented PEvolve, a framework that evolves general rewriting skills from image-generation feedback while keeping model parameters frozen. Through Hierarchical Comparative Reflection, the framework combines within-group comparison and cross-group induction to derive skill edits, and uses Weakness Exploration to test and refine candidate edits. PEvolve improves prompt enhancement with the initial skill library on infographic generation, general-object generation, and knowledge reasoning. Experiments across image generators, including configurations with built-in prompt enhancement enabled and disabled, as well as across PE models and skill initializations, further support the framework's adaptability. Human evaluation on OOD requests that approximate the distribution of real-world user requests also shows that trained skills can transfer rewriting experience acquired from the training tasks to unseen domains. These findings support the use of generation feedback to evolve general rewriting skills.

\clearpage

{\small
\bibliographystyle{plain}
\bibliography{references}
}

\clearpage
\appendix
\raggedbottom
\section{Data Construction and Splits}
\label{app:data_construction}

We organize skill evolution and evaluation around three tasks: infographic generation, general-object generation, and knowledge reasoning. Each task contains 200 training requests, 200 validation requests, and 600 test requests. Training requests provide generation feedback for skill evolution, validation requests determine candidate-library acceptance and final library selection, and test requests are used for final evaluation. These tasks cover information organization, explicit visual constraints, and knowledge-dependent generation.

\paragraph{Infographic Generation.}

The infographic dataset is organized into five content categories: infographics, document images, maps and miscellaneous diagrams, real-world labels, and interfaces and technical diagrams. These requests require the enhancer to organize content, specify text, and express visual hierarchy within a single image. Both the training and validation sets contain 60 infographic requests, 50 document requests, and 30 requests from each of the remaining three categories. The test set contains 169, 146, 95, 95, and 95 requests from these categories, respectively. Table~\ref{tab:appendix_data_categories} reports the category distribution.

\paragraph{General-Object Generation.}

We construct general-object requests using GPT-5.6 Sol and use a selected set of 1,000 requests for the experiments. The requests span 20 content domains, including household objects, tools, animals, plants, food, and transportation. Each request is assigned a primary constraint category---counting, attribute binding, text rendering, negation, or counterintuitive configurations---and may involve additional constraints such as spatial relations, interactions, or multiple actions. The records retain the required entities, constraint annotations, and the central generation difficulty. We partition the selected set using stratification over category, content domain, difficulty, and primary and additional reasoning constraints, while enforcing the target sizes of 200, 200, and 600. Table~\ref{tab:appendix_data_categories} lists the distribution by primary constraint category.

\paragraph{Knowledge Reasoning.}

The knowledge-reasoning task uses WISE~\cite{niu2025wise} as its source benchmark. We use 200 synthesized requests for training and another 200 for validation, and select 600 requests from the original WISE data for testing. The data cover cultural knowledge, biology, chemistry, physics, space, and time, with 26 subcategory labels in our experimental data. Each request is accompanied by reference text describing the expected knowledge-dependent answer. The training and validation sets each contain 79 cultural-knowledge requests, 25 physics requests, and 24 requests from each of the other four domains. The test set contains 240 cultural-knowledge requests and 72 requests from each of the other five domains. This setting evaluates how rewriting skills learned from synthesized requests apply to knowledge-dependent requests from the source benchmark.

\begin{table}[!htbp]
\centering
\small
\renewcommand{\arraystretch}{1.1}
\caption{Data distribution by the primary category of each task.}
\label{tab:appendix_data_categories}
\begin{tabularx}{\textwidth}{@{}Xrrr@{}}
\toprule
Category & Train & Validation & Test \\
\midrule
\multicolumn{4}{@{}l}{\textbf{Infographic generation}} \\
Infographics & 60 & 60 & 169 \\
Document images & 50 & 50 & 146 \\
Maps and miscellaneous diagrams & 30 & 30 & 95 \\
Real-world labels & 30 & 30 & 95 \\
Interfaces and technical diagrams & 30 & 30 & 95 \\
\midrule
\multicolumn{4}{@{}l}{\textbf{General-object generation}} \\
Counting & 86 & 87 & 267 \\
Attribute binding & 28 & 28 & 85 \\
Text rendering & 36 & 36 & 105 \\
Negation & 30 & 30 & 88 \\
Counterintuitive configurations & 20 & 19 & 55 \\
\midrule
\multicolumn{4}{@{}l}{\textbf{Knowledge reasoning}} \\
Cultural knowledge & 79 & 79 & 240 \\
Biology & 24 & 24 & 72 \\
Chemistry & 24 & 24 & 72 \\
Physics & 25 & 25 & 72 \\
Space & 24 & 24 & 72 \\
Time & 24 & 24 & 72 \\
\bottomrule
\end{tabularx}
\end{table}

\paragraph{Out-of-Domain Requests.}

For out-of-domain evaluation, we use OOD data that approximate the distribution of real-world user requests and cover domains outside the three training tasks. We compare the initial skill library with the composed library obtained by merging the validation-best libraries from the three tasks. The human evaluation includes 400 image pairs. These requests are reserved for evaluation and do not participate in skill training, probe construction, skill merging, or library selection. The human evaluation protocol is described in Appendix~\ref{app:evaluation_details}.

\FloatBarrier

\section{Initial Skill Library: Sources and Construction}
\label{app:initial_skill_library}

\paragraph{Library Construction.}

We construct the initial skill library $\mathcal{L}_0$ by decomposing a system prompt into skills. The instructions are organized by rewriting function, with related rules grouped into the same skill. These skills cover general rewriting needs such as scene organization, subject description, text rendering, and visual style. The three tasks in our main experiments share this initial library and evolve it independently using their respective generation feedback.

\paragraph{Skill Organization and Use.}

Each skill is stored as a Markdown document containing a unique identifier, a routing description, and a rule body. The routing description summarizes the skill's scope and rewriting function, while the body provides concrete instructions in natural language, together with examples and cautions where appropriate. Given a user request, the router uses the identifiers and descriptions to select relevant skills. The prompt enhancer then incorporates their descriptions and bodies when rewriting the request. Evolution can edit both routing descriptions and rule bodies.

\paragraph{Alternative Initialization.}

To examine whether PEvolve can improve skills derived from a different system prompt, we construct an alternative initial library by decomposing the official PE prompt of Qwen-Image-2512 into skills. We independently evolve this library for each of the three tasks, using Qwen-Image-2512 as the image generator with its built-in PE disabled. Within each task, we compare prompt enhancement using this initialization with prompt enhancement using its trained counterpart, keeping the routing and rewriting procedure unchanged. Results are reported in Table~\ref{tab:generator_adaptation}.

\section{Automatic and Human Evaluation}
\label{app:evaluation_details}

\paragraph{Automatic Evaluation Setup.}

We use Gemini 3.1 Pro Preview to evaluate generated images against the original user requests, with a task-specific rubric shared across the compared methods. For knowledge reasoning, the judge additionally receives reference text describing the expected answer. The judge returns a score and an accompanying assessment, providing both a scalar evaluation and textual feedback for skill evolution. At test time, each request produces one image. We average image-level scores to obtain task-level metrics and scale the reported scores by 100.

\paragraph{Infographic Generation.}

We evaluate infographic generation along five dimensions: prompt consistency, text accuracy, layout and hierarchy, visual aesthetics, and information richness. Prompt consistency measures whether the requested elements and content are correctly depicted. Text accuracy assesses the correctness and legibility of in-image text. Layout and hierarchy concern the organization, alignment, and visual prioritization of information. Visual aesthetics assesses rendering quality, visual coherence, and artifacts. Information richness measures the substance and useful detail conveyed by the image, accounting for both sparse content and redundant repetition. Each dimension receives an independent continuous score in $[0,1]$, and their equally weighted mean forms the image-level score.

\paragraph{General-Object Generation.}

General-object generation is evaluated on core constraint correctness, object and relation completeness, and prompt adherence and legibility. Core constraint correctness assesses whether the key counting, attribute-binding, text-rendering, negation, or counterintuitive requirement is satisfied. Object and relation completeness checks the presence and distinctness of the required objects, together with their interactions and spatial or causal relations. Prompt adherence and legibility assesses overall compliance with the request and whether the details needed to verify its constraints are sufficiently visible. Each dimension receives a continuous score in $[0,1]$, and the image-level score is their equally weighted mean.

\paragraph{Knowledge Reasoning.}

For knowledge reasoning, the judge determines whether the image correctly realizes the knowledge-dependent answer implied by the original request, using the reference text to establish the expected content. A correct realization receives a score of 1, and an incorrect realization receives 0. The mean binary score gives task accuracy.

\paragraph{Human GSB Evaluation.}

\begin{figure}[H]
\centering
\includegraphics[width=\textwidth]{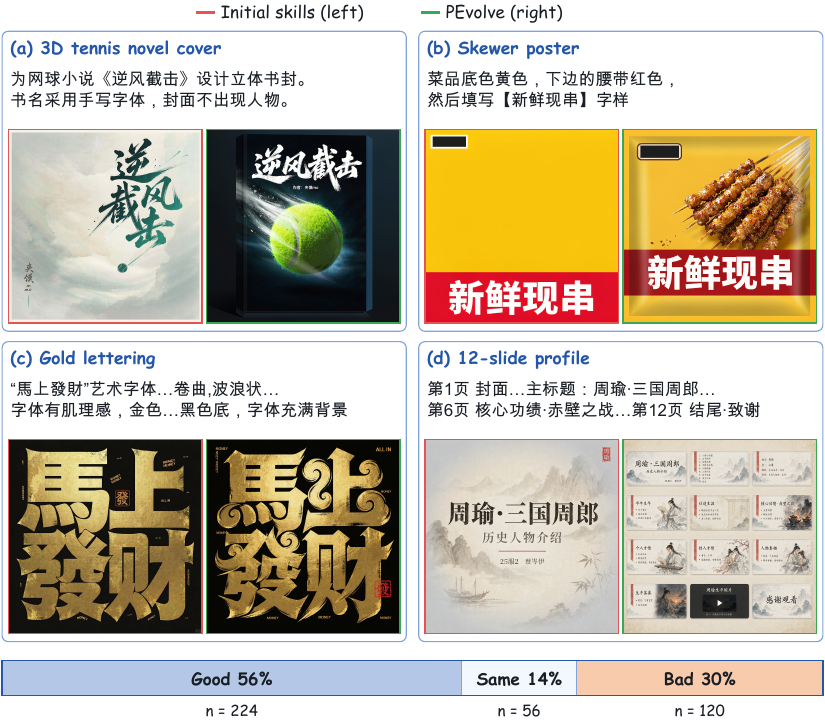}
\caption{Human evaluation on out-of-domain requests. Top: excerpts from the original Chinese prompts and four image pairs for which annotators preferred the output from the merged trained skill library (right) over the initial library (left). Bottom: preference over all 400 image pairs. Good, Same, and Bad indicate that the trained library's output was judged better, comparable, or worse, respectively.}
\label{fig:gsb_human_evaluation}
\end{figure}

We compare the initial skill library with the library merged from the three task-specific trained libraries on 400 OOD image pairs. The requests approximate the distribution of real-world user requests and cover domains outside the three training tasks. Both conditions use the same prompt enhancer, image generator, and generation settings. The annotation interface displays the original request and two images anonymously labeled A and B, while hiding method identities, enhanced prompts, and skill contents. Annotators compare the overall generation outcomes against the original request and select A better, approximately the same, or B better. These choices are mapped to Good, Same, and Bad relative to the initial skill library, indicating that the trained-library output is better, comparable, or worse, respectively. We compute the proportion of each category over all 400 image pairs. Figure~\ref{fig:gsb_human_evaluation} presents the resulting preference distribution.

\section{Core Component Ablation Settings}
\label{app:additional_analyses}

\paragraph{Without Hierarchical Comparative Reflection.}

This variant performs only one generation attempt per user request, producing one rewritten prompt and its corresponding image. Instructions for within-group comparison and cross-group induction are also removed from the analysis prompt. The optimizer proposes candidate skill edits based on the scores, assessments, and visual analyses of the generation results.

\paragraph{Without Weakness Exploration.}

This variant retains Hierarchical Comparative Reflection, with within-group comparison and cross-group induction proposing candidate skill edits. The candidate edits are applied directly to the library, which then undergoes validation to determine whether the update is accepted. This setting removes targeted probe construction, probe evaluation, and iterative refinement of candidate edits.

\paragraph{Shared Settings.}

The ablation experiments cover infographic generation, general-object generation, and knowledge reasoning, using the same data splits, initial skill library, and model configurations as the main experiments. Skill libraries are evolved independently for each task, and the version with the highest validation score is used for final testing. Results are reported in Table~\ref{tab:core_ablation}.

\clearpage
\section{Qualitative Comparisons}
\label{app:qualitative_cases}
\begingroup
\setlength{\intextsep}{4pt}

Figures~\ref{fig:qual_infographic}--\ref{fig:qual_wise} show six cases per task, arranged in two rows and three columns. Each case compares Initial Skills (top) with PEvolve (bottom). Short descriptions summarize the requested content or constraints.

\begin{figure}[H]
\centering
\includegraphics[width=\textwidth]{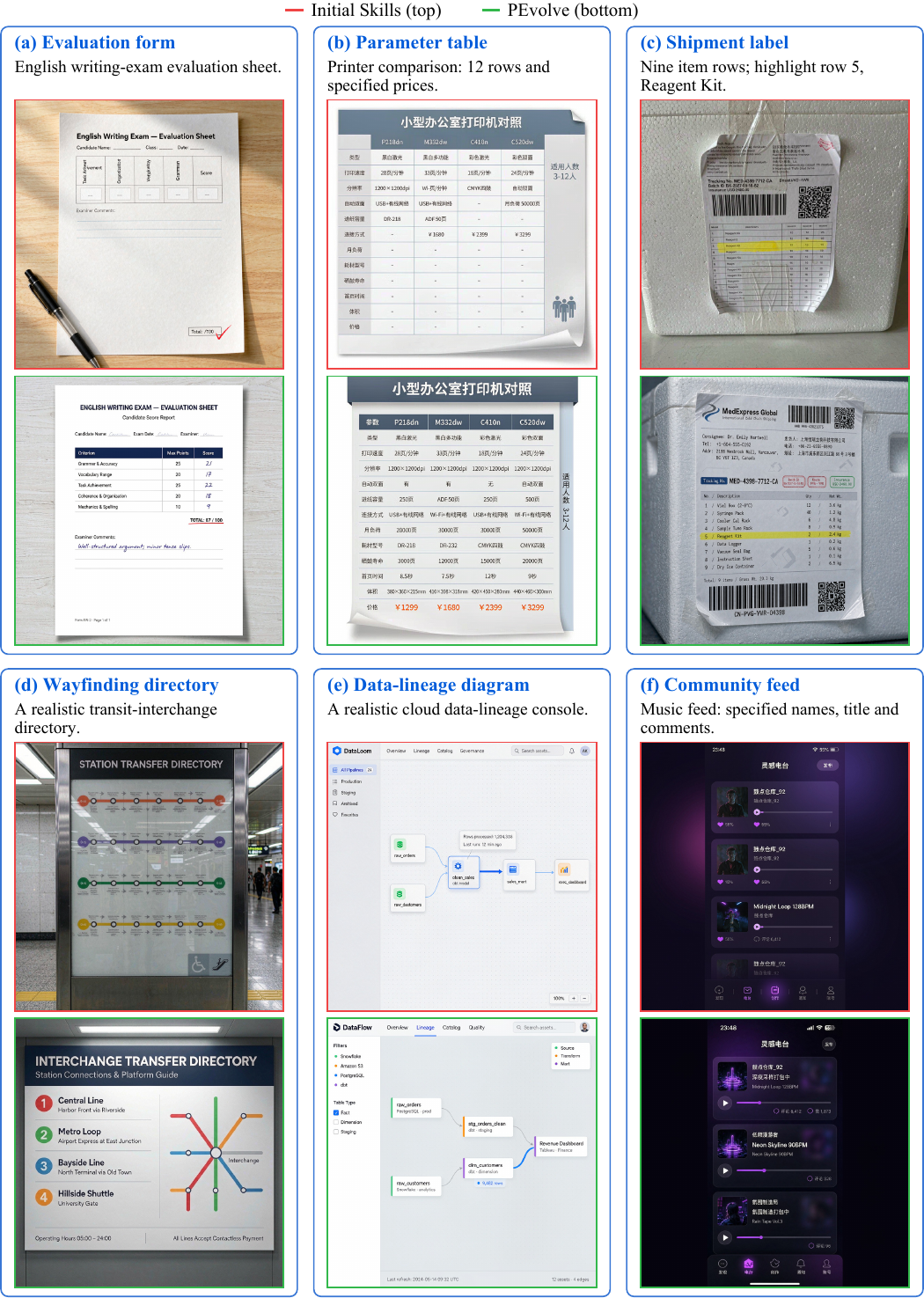}
\caption{Infographic generation: evaluation forms, parameter tables, shipment labels, wayfinding directories, data-lineage diagrams, and community feeds.}
\label{fig:qual_infographic}
\end{figure}

\clearpage
\begin{figure}[H]
\centering
\includegraphics[width=\textwidth]{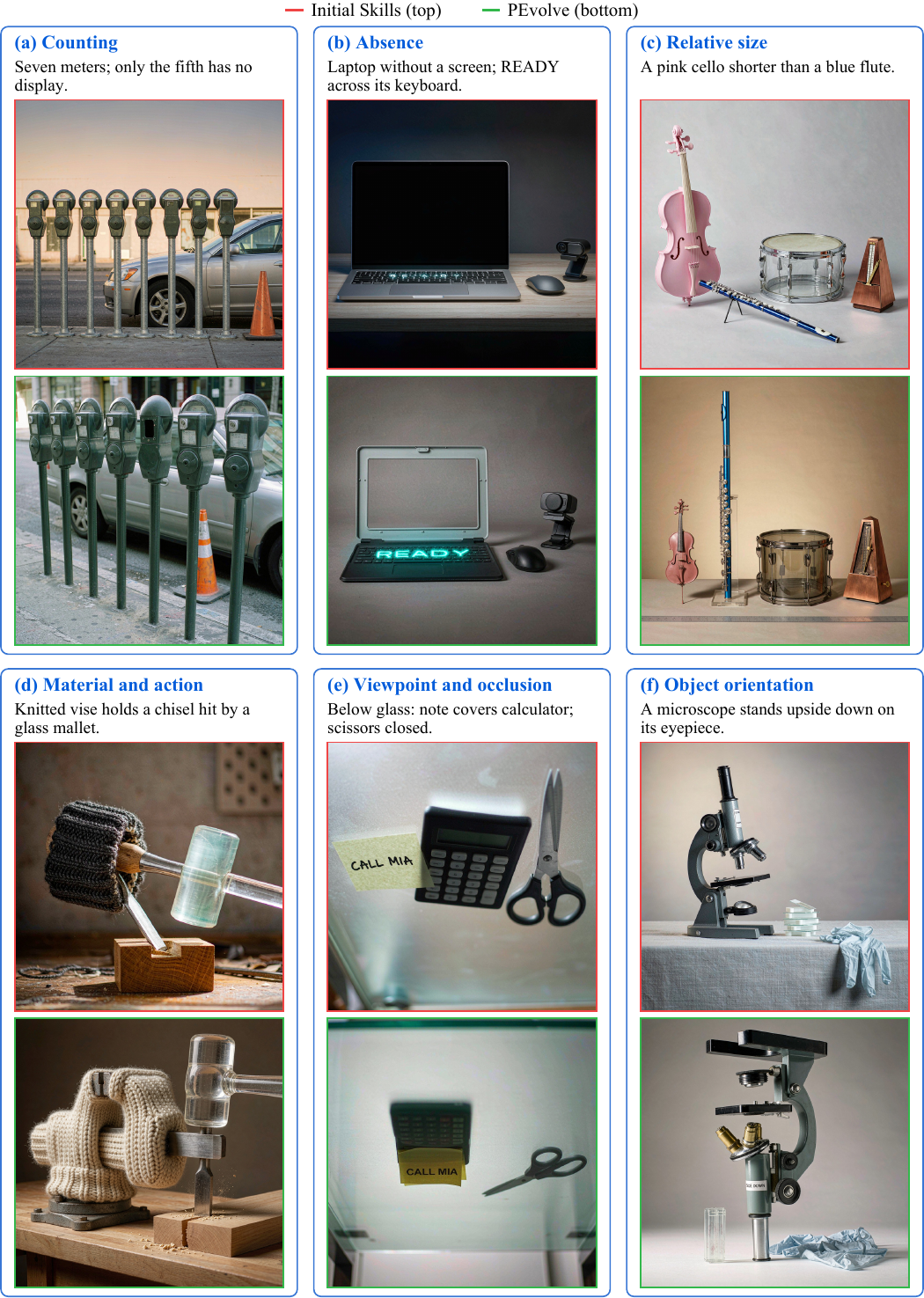}
\caption{General-object generation: counting, absence, relative size, material and action, viewpoint and occlusion, and object orientation.}
\label{fig:qual_general}
\end{figure}

\clearpage
\begin{figure}[H]
\centering
\includegraphics[width=\textwidth]{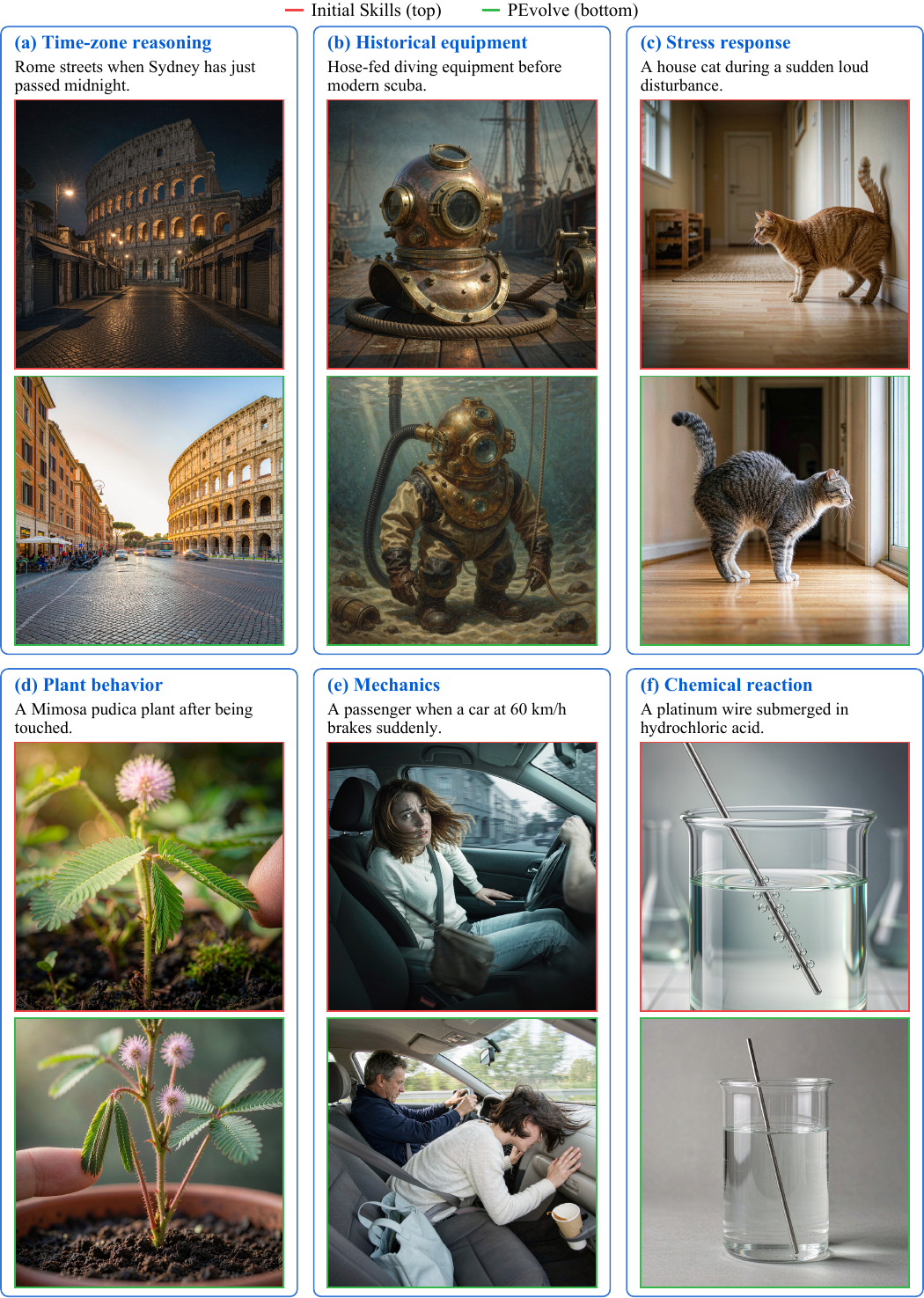}
\caption{Knowledge reasoning: time zones, historical equipment, stress responses, plant behavior, mechanics, and chemical reactions.}
\label{fig:qual_wise}
\end{figure}

\endgroup
\clearpage
\section{Contributors}

\begin{itemize}[leftmargin=*,itemsep=0.7em]
    \item \textbf{Project Supervisors:} Tianyu Pang, Hengshuang Zhao
    \item \textbf{Project Leader:} Zehan Wang
    \item \textbf{Core Contributors:} Jialei Wang, Rui Huang, Yanheng He, Xinyu Yuan
    \item \textbf{Contributors:} Siqi Kou, Ziang Zhang, Xiangyan Liu
\end{itemize}

\end{document}